\documentclass{ifacconf}

\usepackage{graphicx}      
\usepackage{natbib}        
\usepackage[dvipsnames]{xcolor}
\usepackage{amssymb}
\usepackage{multirow}
\usepackage{chemformula}
\usepackage[version=4]{mhchem}
\begin{document}
\begin{frontmatter}

\title{Onboard Health Estimation using Distribution of Relaxation Times for Lithium-ion Batteries} 


\author[First]{Muhammad Aadil Khan} 
\author[First]{Sai Thatipamula} 
\author[First]{Simona Onori} 

\address[First]{Department of Energy Science \& Engineering, Stanford University, Stanford, CA 94305 USA (e-mail: maadilk@stanford.edu, saitv@stanford.edu, sonori@stanford.edu)}

\begin{abstract}                
Real-life batteries tend to experience a range of operating conditions, and undergo degradation due to a combination of both calendar and cycling aging. Onboard health estimation models typically use cycling aging data only, and account for at most one operating condition e.g., temperature, which can limit the accuracy of the models for state-of-health (SOH) estimation. In this paper, we utilize electrochemical impedance spectroscopy (EIS) data from 5 calendar-aged and 17 cycling-aged cells to perform SOH estimation under various operating conditions. The EIS curves are deconvoluted using the distribution of relaxation times (DRT) technique to map them onto a function $\textbf{g}$ which consists of distinct timescales representing different resistances inside the cell. These DRT curves, $\textbf{g}$, are then used as inputs to a long short-term memory (LSTM)-based neural network model for SOH estimation. We validate the model performance by testing it on ten different test sets, and achieve an average RMSPE of 1.69\% across these sets.
\end{abstract}

\end{frontmatter}

\section{Introduction}
The growing popularity of lithium-ion batteries (LIBs) is a direct consequence of their high energy density and long cycling life, which makes them suitable for a large range of applications such as electric vehicles (EVs), consumer electronics, and stationary grid storage. Over the past decade, the material and production costs of LIBs have consistently decreased, promoting their widespread adoption (\cite{knehr2024cost}). In the field, these batteries are exposed to a diverse set of operating conditions, such as discharge rate, temperature, and depth-of-discharge (DOD), which results in complex and unpredictable capacity degradation behaviors. Furthermore, both calendar and cycling aging play a role in battery degradation, which implies that state-of-health (SOH) estimation models using only one type of aging data will not be accurate.

Battery management systems (BMSs) use information from different sensors to make decisions about safe and efficient battery operation. For SOH monitoring, existing models used in the BMS do not provide insight into the evolution of internal degradation mechanisms as the battery ages. Physics-based models (\cite{atlung1979dynamic}, \cite{doyle1993modeling}) are high-fidelity battery models that provide information about internal physicochemical parameters; however, the limited memory and computational power of a BMS make it infeasible to use these models onboard. Instead, empirical models, such as an Equivalent Circuit Model (ECM), are preferred due to their low storage and computational requirements (\cite{tong2013comprehensive}). ECMs provide accurate performance in the operating range for which they are calibrated, but they do not provide any physical insight into internal battery processes. ECMs require extensive lab testing and parameterization at different operating points and aging life to be able to generalize well (\cite{jackey2009parameterization}). This has promoted the adoption of data-driven models for the estimation of battery SOH due to their capability of learning generalizable features from data. These models have low computational requirements during runtime, and by training on sufficient high-quality data, they can effectively augment empirical models onboard the BMS (\cite{xiong2018towards}).

As the battery degrades, the solid-electrolyte interface (SEI) layer grows over the negative electrode, ion diffusion becomes limited due to the loss of active material (LAM), and electrolyte resistance increases due to the irreversible loss of electrolyte in side reactions (\cite{plett2015battery}). Effectively, various coupled processes occur, and it is difficult to isolate their individual impact on battery degradation. However, electrochemical processes tend to occur over a range of frequencies, which makes it possible to observe them through Electrochemical Impedance Spectroscopy (EIS) (\cite{pastor2016identification}). During EIS, when the battery is at equilibrium (steady-state), low-amplitude current or voltage sinusoidal signals are inserted at different frequencies, and the output is measured. The battery response is linear, which helps in identifying impedance information such as electrolyte resistance, charge-transfer resistance, and diffusion resistance from different parts of the EIS curve. In \cite{zhang2020identifying}, complete EIS curves were used with a Gaussian Process Regression (GPR) model for SOH estimation and remaining useful life (RUL) forecasting. The model achieves an $R^2$ value between 0.68 and 0.96 for four different cells. Similarly, \cite{jones2022impedance} used an ensemble of XGBoost models to take EIS curves and battery usage profiles as input and forecast battery capacity amid uneven usage, achieving an error of 8.2\% on the test set without using any historical usage information of the cell. \cite{gasper2022predicting} highlights how high quality EIS measurements are critical for accurate real-time SOH estimation, and an exhaustive search for 2-4 key frequencies should be conducted from a range of frequencies between $10^0$ and $10^3$ Hz. However, key frequencies can change based on operating conditions and battery degradation, which means the SOH model can lose performance.

Typically, EIS is conducted at a limited number of frequency points, and electrochemical processes can overlap if the resolution is not high enough. To overcome this, we use an analytical method known as the Distribution of Relaxation Times (DRT) -- originally proposed by \cite{von1907anal} -- to map EIS curves onto a function $\textbf{g}$. In DRT, peaks are obtained over distinct timescales, which correspond to different internal battery resistances similar to those captured by the EIS curves. Features such as peak height, peak area, and peak location are monitored, and their variation with battery degradation provides useful information for SOH estimation models. \cite{zhang2022degradation} used features extracted from DRT for SOH estimation using GPR and achieved a root mean squared percentage error (RMSPE) of 1-2\%; however, only three cells were used in this work. In \cite{zhao2024hybrid}, a genetic algorithm is combined with a feed-forward neural network for SOH estimation using features extracted from DRT curves via an autoencoder. To the best of our knowledge, none of the existing works have focused on utilizing DRT for onboard health estimation, which presents a significant opportunity to improve battery life and performance through accurate battery monitoring.

In this paper, we develop a data-driven SOH estimation model over a wide operating range that uses complete DRT curves for SOH estimation. By doing so, we eliminate the need to manually extract features from different parts of the curve (with a chance of missing any important information) and allow the model to find the most important features. We first introduce the dataset of both calendar and cycling-aged cells under different operating conditions. Afterwards, the DRT method is introduced, and EIS data is used to obtain DRT curves for all the cells in the dataset. Finally, a long short-term memory (LSTM)-based neural network model is developed that estimates the SOH based on the DRT curves, and the model is validated by testing it on ten different test sets.

\begin{table}[tbhp]
\begin{center}
\caption{Operating conditions of calendar and cycling-aged cells}
\label{tb:op_cond}
\begin{tabular}{c|c|c}
\hline
Operating conditions & Calendar-aged & Cycling-aged \\
\hline
SOC [\%] & 80, 100 & 0-80, 10-90, 0-100 \\
Charge C-rate [1/h] & - & 0.2C, 1C \\
Temperature [$^\circ$C] & 0, 25, 40 & 0, 25, 40\\
Cycles/day & - & 4, 5, 12, 15 \\
\hline
\end{tabular}
\end{center}
\end{table}

\section{Dataset} \label{sec:dataset}
The dataset used in this work consists of 22 cells with NMC/graphite chemistry and a nominal capacity of 5\,Ah. The cells have a voltage range of 4.2–2.5\,V. Five cells are calendar-aged at 80\% and 100\% state-of-charge (SOC) at temperatures of 0$^\circ$C, 25$^\circ$C, and 40$^\circ$C. The remaining 17 cells undergo cycling aging using a constant current-constant voltage (CCCV) charge/CC discharge cycling protocol with charging C-rate of 0.2C and 1C, and SOC ranges of 0–80\%, 10–90\%, and 0–100\% at the same three temperatures (see Table\,\ref{tb:op_cond}). Cell characterization is conducted at 25$^\circ$C using a capacity test at 1C C-rate and an EIS test at 0\%, 25\%, 50\%, 75\%, and 100\% SOC. These tests are performed periodically at 0, 10, 20, 40, and 90 days of aging.
\begin{figure}
\begin{center}
\includegraphics[width=8.4cm]{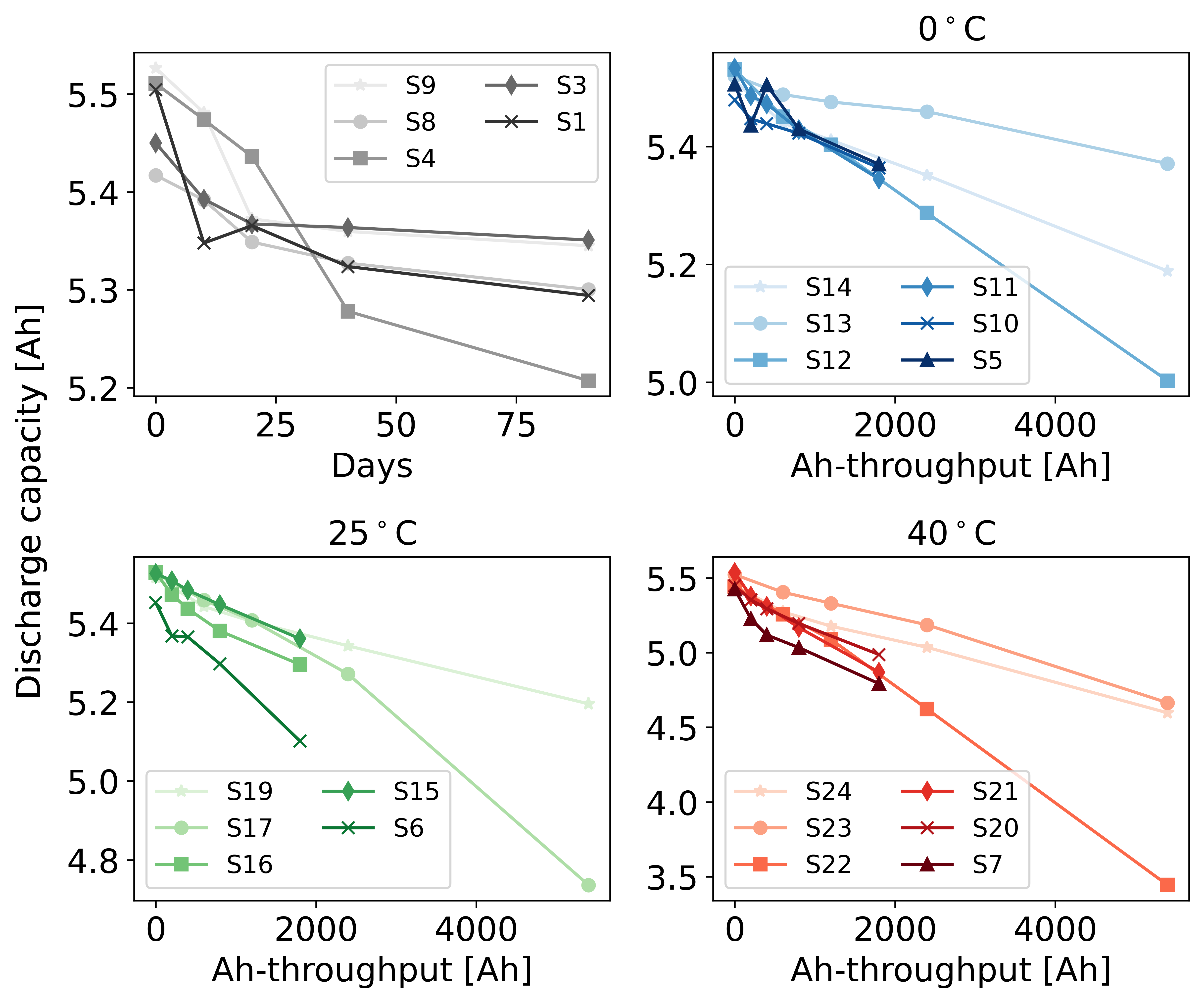}    
\caption{Discharge capacity of 22 cells in the dataset as a function of days for calendar-aged cells (top-left), and Ah-throughput for cycling-aged cells (top-right and bottom). Cell S4, S12, S17, and S22 show maximum degradation in each panel.}
\label{fig:capacity_plots}
\end{center}
\end{figure}

As shown in Fig.\,\ref{fig:capacity_plots}, discharge capacity data have a lot of variation among different cells due to varying operating conditions. Calendar-aged cells do not degrade below 5.2\,Ah, with cells S3 and S9 experiencing the least amount of degradation at the end of 90 days. Cell S22 degrades to 65\% of its nominal capacity because of high temperature and high Ah-throughput. Furthermore, cells aged at 40$^\circ$C exhibit higher aging compared to cells aged at other temperatures; however, some cells also overlap in their degradation trajectories despite being aged at different temperatures, such as cells S14 and S19 (both end up at approximately 5.2\,Ah). This suggests that, apart from temperature, the number of cycles and the SOC range also impact cell aging. Cell S6, which is cycled between 0-100\% SOC at 0.2C, ages faster than Cell S23, which is cycled between 0-80\% SOC at 1C, despite being at a lower temperature. A 0-80\% SOC range is equivalent to a reduced voltage range of 2.5-3.96\,V, which results in a reduced rate of degradation (\cite{aiken2022li}).
\begin{figure*}
\begin{center}
\includegraphics[width=18cm]{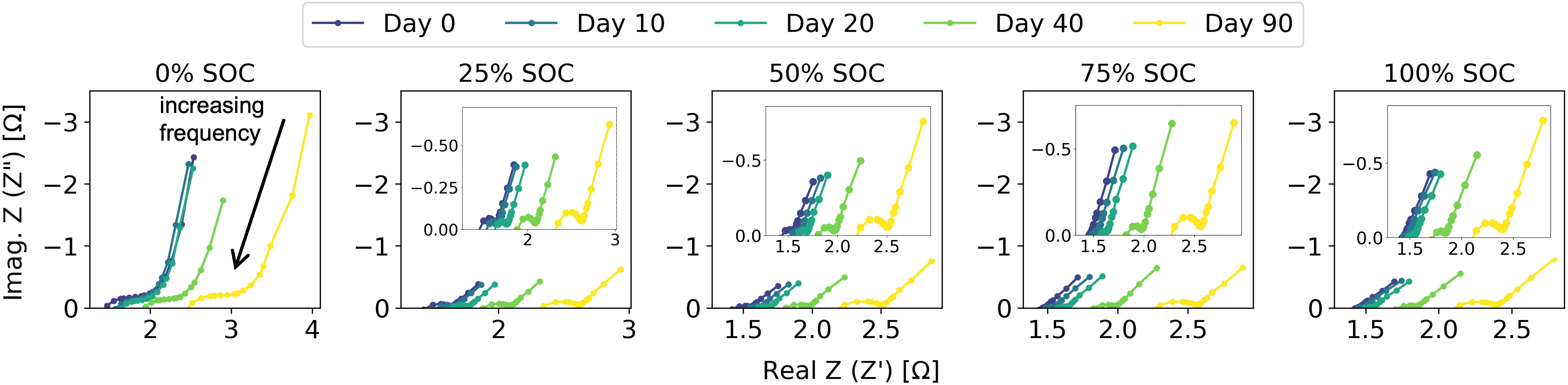}    
\caption{Example EIS curves for Cell S22 at five different SOCs, and 0, 10, 20, 40 and 90 days of aging. Magnitude of impedance is large at 0\% SOC, but relatively small at other SOCs. Black arrow indicates the direction of increasing frequency for all EIS curves.}
\label{fig:eis}
\end{center}
\end{figure*}

\section{Methodology}
\begin{figure*}[thbp]
\begin{center}
\includegraphics[width=14.5cm]{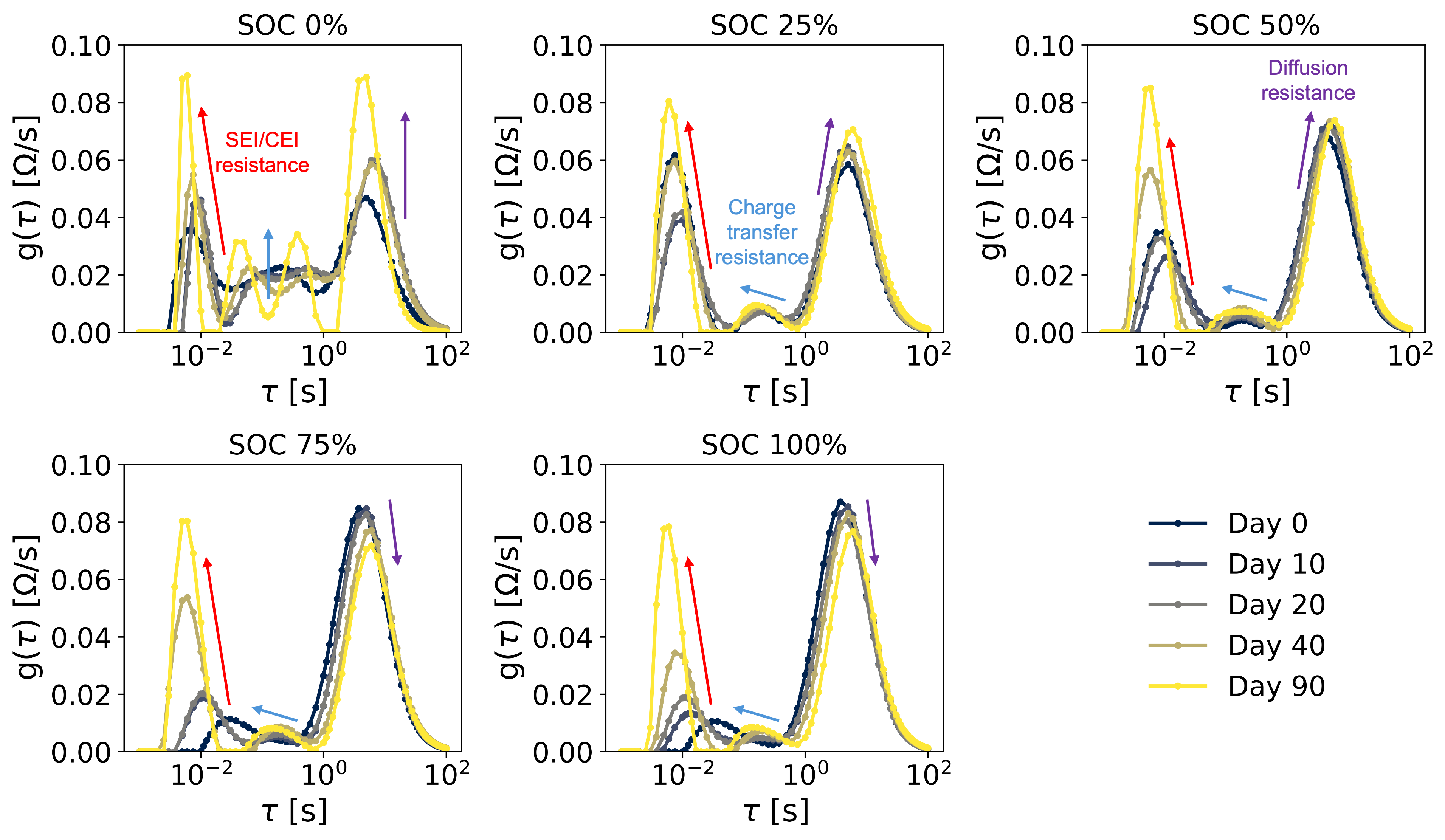}    
\caption{Example DRT curves for Cell S22 at five SOCs and 0, 10, 20, 40 and 90 days of aging. With aging, SEI/CEI resistance peak ($\tau \approx10^{-2}$s) shows an increase in height and shifts to the left. Charge transfer resistance ($10^{-2} < \tau < 10^0$s) shows two peaks for 0\% SOC which increase in height while for other SOCs, it shows a small increase in height and shifts to the left. Diffusion resistance peak ($\tau > 10^0$s) increases in height for 0\%, 25\% and 50\% SOC, but it shows a small decrease in height at 75\% and 100\% SOC.}
\label{fig:drt_curves_all}
\end{center}
\end{figure*}
\subsection{Distribution of Relaxation Times}
The EIS spectrum is a function of both battery aging and SOC, as shown in Fig.\,\ref{fig:eis}. At 0\% SOC, the cathode is fully lithiated while the anode is fully delithiated. Ions have less space to diffuse out of the cathode, resulting in a relatively large magnitude of both the real impedance ($Z'$) and the imaginary impedance ($Z''$) compared to their magnitudes at other SOCs. The shape of the EIS curve varies and changes with both SOC and aging, making it difficult to directly identify and use information about different degradation mechanisms. The DRT method deconvolves the EIS spectrum by modeling it as a sum of high-frequency resistance $R_0$, and the contribution of different time constants through an integral term (\cite{paul2021computation}). The impedance spectrum $Z(\omega) \in \mathbb{R}^m$ is given by
\begin{equation} \label{eq:DRT_eq}
    Z(\omega) = R_0 + R_p \int_0^\infty \frac{g(\tau)}{1 + i \omega \tau} d\tau
\end{equation}
where $\omega \in \mathbb{R}^m$ is the frequency, $\tau \in \mathbb{R}^n$ is the time constant, $R_0 \in \mathbb{R}$ is the high-frequency resistance, $R_p \in \mathbb{R}$ is the polarization resistance, and $g(\tau)$ is the continuous DRT function. To solve equation\,(\ref{eq:DRT_eq}), $Z(\omega)$ is decomposed into its real and imaginary parts, and rewritten numerically as
\begin{equation} \label{eq:Z_real}
    Z'(\omega_m) = R_0 + R_p \sum_{n=1}^N \frac{g_n}{1 + \omega_m^2 \tau_m^2} \delta \tau_n
\end{equation}
\begin{equation} \label{eq:Z_imag}
    Z''(\omega_m) = -R_p \sum_{n=1}^N \frac{\omega_m \tau_n g_n}{1 + \omega_m^2 \tau_m^2} \delta \tau_n
\end{equation}
A linear system of equations can be set up of the form $\textbf{A}\textbf{g}=\textbf{Z}$ where $\textbf{A} \in \mathbb{R}^{m \times n}$ is a matrix that contains the sum of $R_0$ and RC elements, $\textbf{Z} \in \mathbb{R}^{m}$ is a vector of impedances at different frequencies $\omega_m$, and $\textbf{g} \in \mathbb{R}^n$ is the unknown DRT function. Either equation\,(\ref{eq:Z_real}) or (\ref{eq:Z_imag}) or both combined can be used to solve for $\textbf{g}$. In this work, we use equation\,(\ref{eq:Z_real}) to solve for $\textbf{g}$; however, directly solving for $\textbf{g}$ results in an infeasible solution due to the high condition number of matrix $\textbf{A}$. Effectively, $\textbf{A}\textbf{g}=\textbf{Z}$ is an ill-posed problem with no unique solutions. One approach used to overcome this is to use Tikhonov regularization (\cite{paul2021computation}); then, $\textbf{g}$ is given by
\begin{equation} \label{eq:g_solve}
   \textbf{g} = (\textbf{A}^T \textbf{A} + \lambda I)^{-1} \textbf{A}^T \textbf{Z}
\end{equation}
where $\lambda$ is the regularization parameter. When solving equation\,(\ref{eq:g_solve}), a positivity constraint is imposed on $\textbf{g}$ to ensure that the results are physically plausible since $\textbf{g}$ has units of resistance per second. The optimal value of $\lambda$ is obtained using the L-curve method (\cite{engl1994using}), by finding the point where both the residual norm $||\textbf{A} \hat{\textbf{g}} - \textbf{Z}||_2$ and the solution norm $||\hat{\textbf{g}}||_2$ are minimized. Fig.\,\ref{fig:drt_curves_all} shows the DRT curves for cell S22 as a function of SOC and aging. Left peak shows the evolution of SEI/CEI resistance with aging and generally has an increasing trend for all SOCs. In the middle, peaks correspond to charge-transfer resistance and show little variation with aging apart from at 0\% SOC where two peaks are observed. Right peak corresponds to diffusion resistance and shows an increasing trend for 0\% to 50\% SOC, but a decreasing trend at 75\% and 100\% SOC. One reason for this behavior can be the enhanced mobility of lithium-ions at high SOC due to a uniform concentration gradient which was absent from low to middle SOC values.

\subsection{Model Architecture} \label{subsec:model_arch}
\begin{figure*}[thbp]
\begin{center}
\includegraphics[width=17cm]{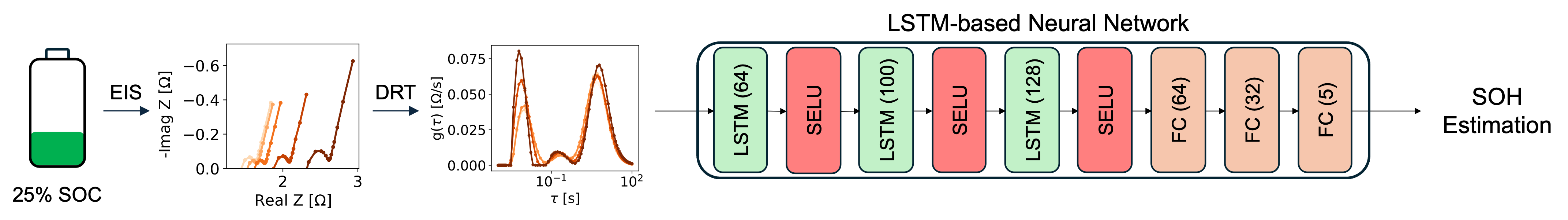}    
\caption{SOH pipeline representing flow of impedance data and architecture of the data-driven model. Data from EIS at 25\% SOC is transformed into DRT curves, which are used as input to the LSTM-based model. The model has three LSTM layers with SELU activation followed by three fully-connected layers. On the output, the model provides SOH estimation at different days.} 
\label{fig:model}
\end{center}
\end{figure*}

The model used in this work is a neural network based on an LSTM, and the estimation problem is set up as a sequence-to-sequence problem. The SOH estimation pipeline and the model architecture are shown in Fig.\,\ref{fig:model}. Three LSTM layers enable the model to extract information about short-term and long-term dependencies present in the data. After each LSTM layer, a Scaled Exponential Linear Unit (SELU) activation is applied to induce non-linearity and allow the model to learn the intricate behavior of different resistances through the DRT curve. SELU also has a self-normalizing quality, which helps in stable training of the model by avoiding vanishing or exploding gradients. 
Finally, three fully-connected (FC) layers are added for the final regression task with linear activation. In this architecture, the depth of the neural network is increased to ensure it can learn the complex evolution of degradation mechanisms from DRT curves to accurately estimate the SOH.
The model consists of 210,645 trainable parameters, and the model size is less than 1\,MB, making it a good candidate for onboard applications. For model training, we use an Adam optimizer with an initial learning rate of 0.001 and a mean squared error (MSE) loss function. A variable learning rate is used, which changes by a factor of 0.5 whenever the model performance on the validation set hits a plateau.

\section{Simulation Results and Discussion}
Since we are interested in onboard health estimation, we first attempt to find the appropriate SOC for EIS measurement and DRT calculation, specifically with the goal of SOH estimation. This requirement stems from an implementation standpoint: EIS measurements at a particular SOC should be used for SOH estimation such that experimental resource requirements are minimized. To extend battery life, it is generally recommended to avoid fully discharging (SOC $\rightarrow$ 0\%) and fully charging (SOC $\rightarrow$ 100\%) a battery. Hence, from the remaining DRT curves at 25\%, 50\% and 75\% SOC, we select DRT curves at 25\% SOC for the SOH estimation model. From Fig.\,\ref{fig:drt_curves_all}, it is evident that DRT curves obtained from EIS data collected at 0-50\% SOC show more variation in terms of peak behaviors, which can be helpful for the SOH model to learn useful features. This behavior is consistently seen in DRT curves calculated for other cells in the dataset as well. Furthermore, initial studies conducted using DRT curves obtained at various SOCs show that DRT curves at 25\% SOC gave the best overall performance for SOH estimation.

Three different test set categories are created to validate the model performance: 1) balanced, 2) temperature-based, and 3) randomized. Furthermore, the dataset split is also varied to analyze the impact of the size of the training set on model performance. The LSTM model performance is compared against a linear regression model using root mean squared error (RMSE) and root mean squared percentage error (RMSPE) given by:
\begin{equation}
    \text{RMSE} = \sqrt{\frac{1}{K} \sum_{i=1}^K (y_i - \hat{y_i})^2}
\end{equation}
\begin{equation}
    \text{RMSPE} = \sqrt{\frac{1}{K} \sum_{i=1}^K \left(\frac{y_i - \hat{y_i}}{y_i}\right)^2} \times 100
\end{equation}
where $i=1,...,K$ is the number of SOH points, $y_i$ is the true SOH, and $\hat{y_i}$ is the estimated SOH. 

\begin{figure}[thbp]
\begin{center}
\includegraphics[width=8cm]{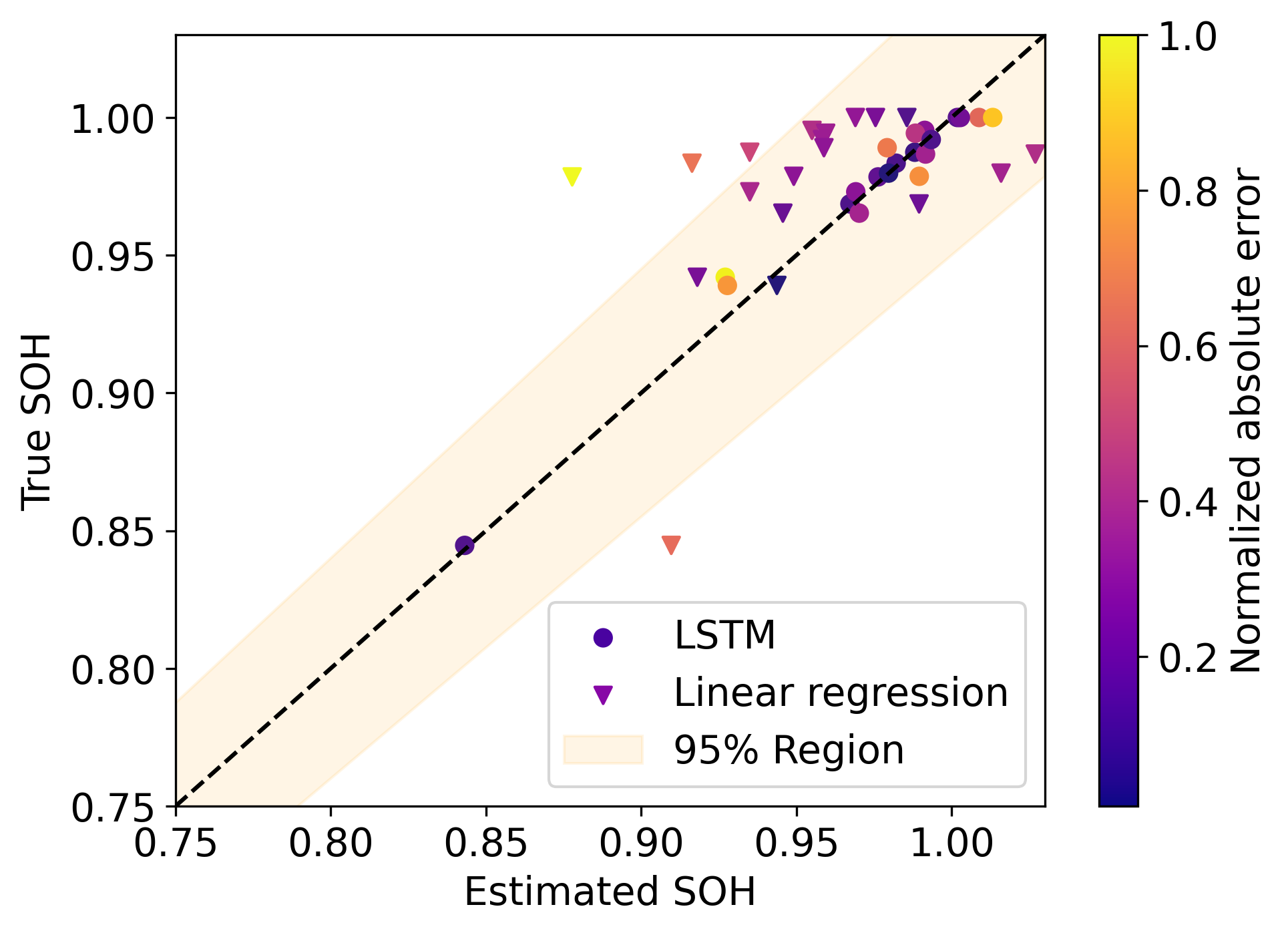}    
\caption{Estimation results of test set 1 consisting of cells S8, S13, S19, and S23. LSTM has good performance in comparison to the experimental data with an RMSPE of 0.7129\%. Linear regression fails to capture the SOH behavior, and underestimates the SOH for majority of the test points.} 
\label{fig:balanced}
\end{center}
\end{figure}

The balanced test sets consist of one cell from each cycling temperature, and additionally, one calendar-aged cell. Fig.\,\ref{fig:balanced} shows the performance of LSTM and linear regression models on the balanced test set consisting of cells S8, S13, S19, and S23. LSTM shows superior performance on all four cells with an RMSE of 0.0069\,Ah and an RMSPE of 0.7129\%, as given in Table\,\ref{tb:model_perf}. The linear regression model fails to capture the SOH behavior for most of the cells resulting in an overall RMSE and RMSPE of 0.0426\,Ah and 4.4215\%, respectively. From Table\,\ref{tb:model_perf}, for other balanced test sets, the LSTM model still has good performance with a maximum RMSPE of 1.3343\%.

\begin{figure}[thbp]
\begin{center}
\includegraphics[width=8cm]{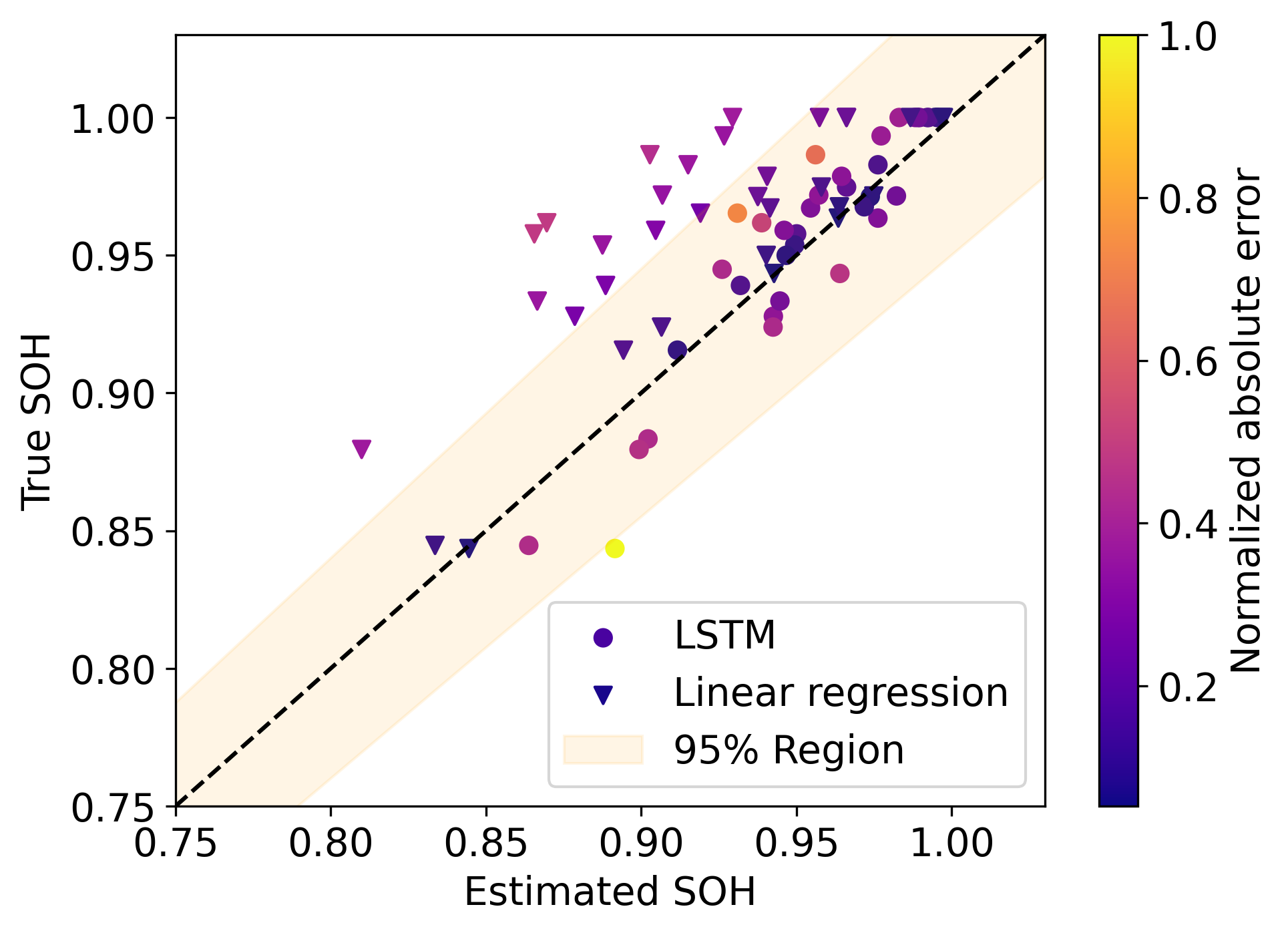}    
\caption{Estimation results of test set 5 consisting of cells S1, S4, S7, S20, S21, S23, and S24. The model is trained on cells only at 0$^\circ$C and 25$^\circ$C, but it performs well on the test set with an RMSE of 0.0167\,Ah and an RMSPE of 1.8221\%. Linear regression model is not able to generalize, and consistently underestimates the SOH.} 
\label{fig:temperature_based}
\end{center}
\end{figure}

For temperature-based test sets, all cells (calendar and cycling) at one temperature are combined into a single test set, and the model is only trained on cells from the other two temperatures. In Fig.\,\ref{fig:temperature_based}, the results are shown for all the cells at 40$^\circ$C (test set 5). As shown in Table\,\ref{tb:model_perf}, despite only training on data from 0$^\circ$C and 25$^\circ$C, the LSTM model performs well with an RMSE of 0.0167\,Ah and an RMSPE of 1.8221\%. Even when tested on cells at 0$^\circ$C and 25$^\circ$C, the RMSPE remains around 2\%. Specifically, for test set 5 at 40$^\circ$C, it is noteworthy that these cells experience significantly higher degradation compared to cells at other temperatures. This demonstrates that the model can extract useful features common to cells at all three temperatures through DRT curves. For onboard scenarios, including high-degradation cells in the training data can help the model perform better by allowing it to learn from a broader range of degradation trajectories. In contrast, the linear regression model struggles with this test set category, with a maximum RMSPE of 6.7536\% due to its poor generalization.

\begin{figure}[thbp]
\begin{center}
\includegraphics[width=8cm]{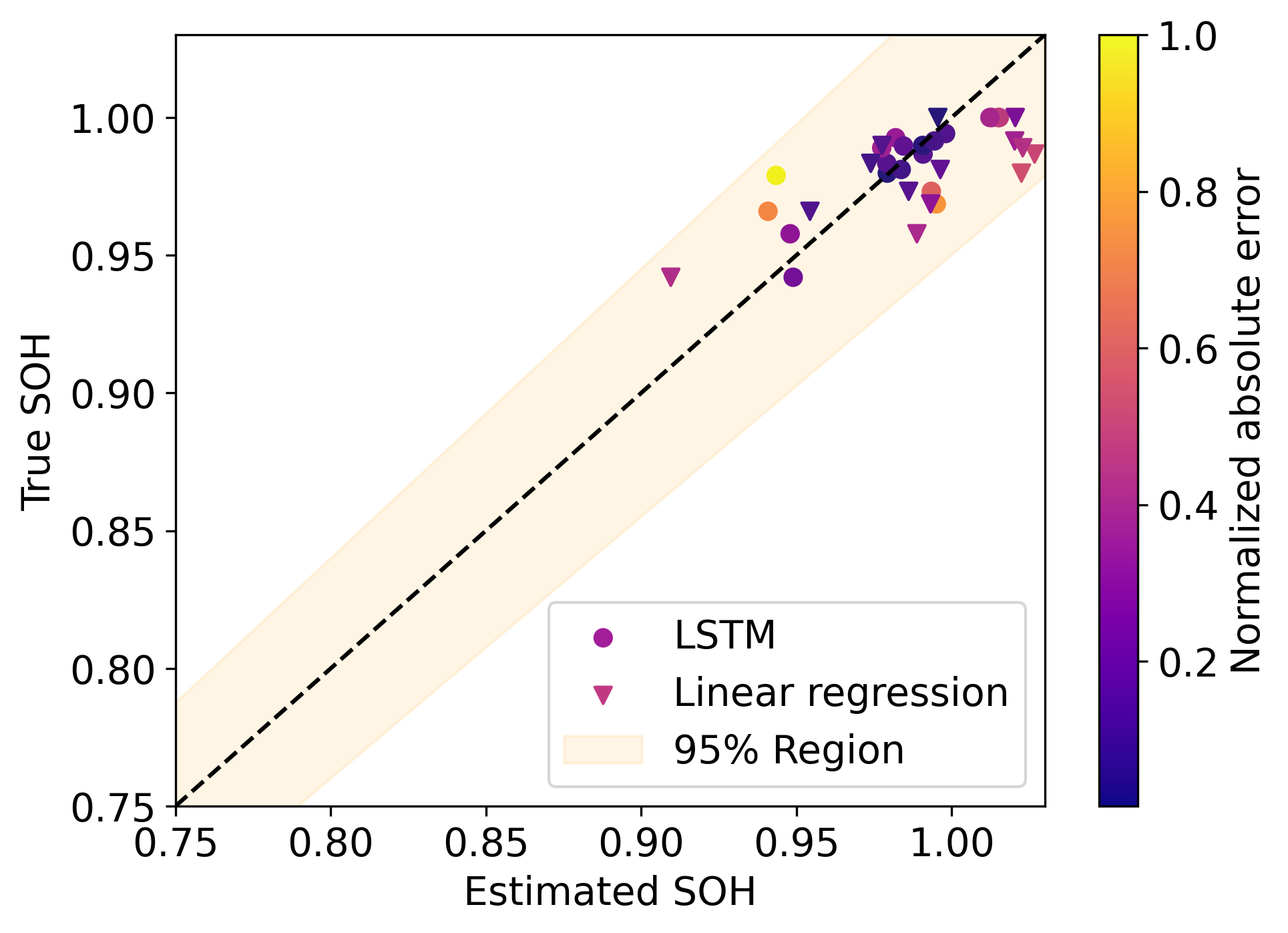}    
\caption{Estimation results of test set 8 consisting of cells S10, S19, S16, and S11. The model achieves an RMSPE of 1.4927\% on the randomized test set despite the cells being operated under different conditions. Linear regression model has slight improvement in performance on this test set as compared to other ones with an RMSPE of 3.7283\%.} 
\label{fig:randomized}
\end{center}
\end{figure}

The last test set category consists of a collection of randomly selected cells under different operating conditions. This test set can be considered as a reasonable reference for real-life applications where operating conditions can vary randomly, with little to no control over temperature, discharging scenarios, and rest times (calendar aging). As illustrated in Fig.\,\ref{fig:randomized}, for test set 8, the LSTM model efficiently captures the SOH trajectories for all the cells, achieving an RMSE of 0.0146\,Ah and an RMSPE of 1.4927\%, as shown in Table\,\ref{tb:model_perf}. For test sets 7 and 9, the performance is also good with an RMSPE of approximately 1.6\% and 2.4\%, respectively. Lastly, the model is trained and tested on half of the cells randomly selected as given in test set 10 in Table\,\ref{tb:model_perf}. The model maintains good performance with an RMSPE of 2.1876\%. With 22 cells, this dataset can be categorized as medium to large, and the LSTM model extracts useful features from the training data and shows good generalization across a range of test sets.

\begin{table*}[tbhp]
\begin{center}
\caption{Model performance on ten different test sets}\label{tb:model_perf}
\begin{tabular}{ccccccc}
 & & & \multicolumn{2}{c}{LSTM} & \multicolumn{2}{c}{Linear Regression} \\
Category & Set \# & Test cells & RMSE [Ah] & RMSPE [\%] & RMSE [Ah] & RMSPE [\%] \\
\hline
         & 1 & S8, S13, S19, S23 & \textbf{0.0069} & \textbf{0.7129} & 0.0426 & 4.4215 \\
Balanced & 2 & S4, S11, S16, S20 & 0.0099 & 1.0134 & 0.0404 & 4.1251 \\
         & 3 & S9, S10, S15, S21 & 0.0127 & 1.3343 & 0.0274 & 2.8287 \\
\hline
                  & 4 & S6, S8, S15, S16, S17, S19 & 0.0204 & 2.2960 & 0.0508 & 5.2733 \\
Temperature-based & 5 & S1, S4, S7, S20, S21, S23, S24 & \textbf{0.0167} & \textbf{1.8221} & 0.0635 & 6.7536 \\ 
                  & 6 & S3, S5, S9, S10, S11, S12, S13, S14 & 0.0195 & 2.0035 & 0.0457 & 4.6736 \\
\hline
           & 7 & S12, S9, S23, S4 & 0.0158 & 1.6608 & 0.0418 & 4.3232 \\
	   & 8 & S10, S19, S16, S11 & \textbf{0.0146} & \textbf{1.4927} & 0.0366 & 3.7283 \\ 
Randomized & 9 & S23, S6, S7, S16 & 0.0217 & 2.4347 & 0.0299 & 3.3394 \\
          &  \multirow{2}{*}{10} & S9, S3, S21, S20 & \multirow{2}{*}{0.0200} & \multirow{2}{*}{2.1876} & \multirow{2}{*}{0.0916} & \multirow{2}{*}{9.3847} \\
           & & S4, S1, S5, S12, S17, S11, S14 & & & & \\
\hline
\end{tabular}
\end{center}
\end{table*}

\section{Conclusion} \label{sec:conclusion}
To conclude, in this paper, we utilized a new dataset consisting of calendar and cycling-aged cells at various operating conditions for SOH estimation. Capacity and EIS tests were periodically conducted for cell characterization, and EIS data was transformed into DRT curves using Tikhonov regularization. The DRT curves were used as input to an LSTM-based neural network model, and the robustness of the model was validated by testing the model on ten different test sets with the best testing error obtained to be 0.7129\%. Overall, the model achieves an average test RMSPE of 1.69\% across all the test sets. We also observed that including cells with higher degradation in the training data helps the model perform better since it can to generalize to a wider range of SOH trajectories. With impedance measurements available onboard the BMS, this model can be an effective augmentation to the existing onboard SOH models. In the future, we plan to extend this work to analyze the impact of individual degradation modes from the DRT curve on SOH, and using it to predict failure of lithium-ion batteries.
\begin{ack}
The authors would like to acknowledge Nuvoton Technology Corporation Japan (NTCJ) for providing funding and the dataset through Stanford SystemX Alliance.
\end{ack}

\bibliography{ifacconf}             
                                                   







\end{document}